\documentclass[
]{ceurart}

\usepackage{listings}
\usepackage{algorithm}
\usepackage{algpseudocode}
\usepackage{xurl}  
\usepackage{enumitem}
\setlist[itemize]{topsep=0.25ex, itemsep=0.15ex, parsep=0pt, partopsep=0pt}
\newcommand{\blockpara}[1]{\par\vspace{0.35\baselineskip}\noindent\textbf{#1}\par\vspace{0.15\baselineskip}}
\newcommand{\defitem}[2]{\noindent\textbf{#1}\ #2\par\vspace{0.1\baselineskip}}
\newcommand{\SupplementRepo}{\href{https://github.com/park-jsdev/classroom-feedback-mediator}{\nolinkurl{github.com/park-jsdev/classroom-feedback-mediator}}}
\begin{document}

\copyrightyear{2026}
\copyrightclause{Copyright for this paper by its authors.
  Use permitted under Creative Commons License Attribution 4.0
  International (CC BY 4.0).}

\conference{HAI-Agency: Workshop on Orchestrating Human and AI Agency for Proactive and Reflective Learning, co-located with the 27th International Conference on Artificial Intelligence in Education (AIED 2026), Seoul, Korea, June 27 - July 3, 2026.}


\title{Surfacing Isolated Learners with Outcome-Independent Mediation of Feedback between Teachers and Students Using AI}


\author[1,2]{Junsoo Park}[
orcid=0009-0003-4450-0103,
email=jpark3232@gatech.edu,
]
\author[2]{Youssef Medhat}[
email=ymedhat3@gatech.edu,
]
\author[2]{Htet Phyo Wai}[
email=hwai6@gatech.edu,
]
\author[2]{Ploy Thajchayapong}[
orcid=0009-0000-5993-1094,
email=ploy@gatech.edu,
]
\author[2]{Ashok K. Goel}[
orcid=0000-0003-4043-0614,
email=ashok.goel@cc.gatech.edu,
]



\cortext[1]{Corresponding author.}
\address[2]{Georgia Institute Of Technology, North Avenue, Atlanta, GA 30332, USA}

\begin{abstract}
AI-augmented classrooms generate rich teacher and student feedback before graded outcomes become available, yet these signals can be difficult to translate into timely instructional decisions. We propose an \emph{interpretable decision layer}: a transparent mechanism that ranks course topics requiring attention without using grades or post-hoc outcome labels. The approach combines three signals: student learning difficulty prevalence, disagreement between learner self-reports and observed difficulties, and unresolved teacher concerns. The output is a ranked set of topic priorities with per-topic decision records explaining each ranking. In one graduate CS course offering ($n=5$ instructor interviews; $n=279$ survey responses), prioritized topics aligned with instructor concerns (top-5 overlap 3/5; Spearman $\rho=0.80$) and student-reported topic difficulty ($\rho=0.46$, $p=.048$). Multi-signal integration also surfaced learners not identified through individual signal sources alone (AUC $=0.96$ vs.\ $0.91$ for gap prevalence alone). Reflective thinking, help-seeking, and self-efficacy provided additional evidence that student behavioral signals align with learning-related constructs. While preliminary, these findings suggest that transparent coordination mechanisms may help support human--AI co-agency when feedback is incomplete.
\end{abstract}


\maketitle

\section{Introduction}

AI teaching assistants are increasingly used in real classrooms. Teachers and learners generate rich sources of information: interviews, surveys, and interaction logs, but these signals are difficult to combine into clear instructional actions \cite{hattie2007power,tomlinson1999differentiated,bernacki2021systematic,plass2020toward}. Graded outcomes often arrive too late to support decisions during learning. We therefore ask: \emph{how can teacher and learner feedback be coordinated when outcome data are unavailable?} We propose an \emph{interpretable decision layer}: a transparent mechanism that combines teacher and student feedback to rank course topics needing attention. Rather than using grades or outcome labels, the approach relies on three signals: i) how common a learning difficulty is among students (gap prevalence), ii) disagreement between learner self-reports and observed difficulties, and iii) unresolved teacher concerns. These signals are combined into ranked topic priorities together with a \emph{decision record} that explains why a topic received its rank. The system organizes teacher, student, and researcher roles around this shared decision process. In one graduate CS course offering ($n\!=\!5$ instructor interviews; $n\!=\!279$ survey responses), prioritized topics aligned with instructor concerns and student-reported difficulty (Section~\ref{sec:validation}). Reflective thinking, help-seeking, and self-efficacy aligned with topic understanding as complementary evidence of learning-related behavior~\cite{kember2000rtq,bandura2006guide}, while combining multiple signals surfaced learners missed by single-source approaches.

\blockpara{Definitions.}
\defitem{Topic.}{A knowledge-graph node representing a course concept (e.g., planning, analogical reasoning); this offering contains 54 topics and 47 prerequisite links (Appendix~\ref{app:kg}).}
\defitem{Gap prevalence ($R_t$).}{The fraction of students identified as having difficulty with topic~$t$.}
\defitem{Survey disagreement ($D_t$).}{Difference between observed topic difficulty and student self-reported difficulty: $|R_t-s_t|$ when survey difficulty $s_t$ is available (Eq.~\ref{eq:d_t}).}
\defitem{Teacher friction ($F$).}{A summary measure of unresolved instructor concerns derived from coded interview themes (Appendix~\ref{app:teacher}).}
\defitem{Mediation layer.}{Combines $(R_t,D_t,F)$ into a topic priority score $P_t$ and ranking (Eq.~\ref{eq:p_t}).}
\defitem{Decision record.}{Per-topic inputs, score, rank, and diagnostics. Figure~\ref{fig:arch} uses the synonymous label ``Decision Trace.''}
\defitem{Isolated learner.}{A learner not identified by any single signal source, but surfaced when multiple signals are considered together.}
\vspace{-0.1\baselineskip}

\blockpara{Research question and hypotheses.}
\defitem{RQ:}{Can a transparent decision layer combine teacher and student feedback to identify useful support priorities without relying on grades or other outcome labels?}
\defitem{H1.}{The mediation layer produces interpretable topic priorities that remain similar across reasonable weight choices.}
\defitem{H2.}{Student behavioral signals such as reflective thinking and help seeking are associated with learning-related measures at the cohort level.}
\defitem{H3.}{Combining multiple signals identifies learners who would not be identified by any individual signal alone.}
\vspace{-0.1\baselineskip}

\section{Related Work}

\paragraph{Teacher orchestration and reflection.}
Prior work highlights the need for teacher \emph{orchestration} in AI-enhanced classrooms and for interfaces that help instructors interpret learner activity and make instructional decisions~\cite{holstein2019complementarity,hershkovitz2024instructors,cai2025aipoweredreflection}. Open and inspectable learner models further emphasize the importance of transparency in educational AI systems~\cite{bull2016smili}. Our work builds on these ideas by focusing on an interpretable coordination process in which topic priorities can be traced back to their contributing signals. Unlike approaches centered on post-hoc analytics or dashboard reflection alone, we consider how teacher and student perspectives can be combined during instruction when outcome measures are not yet available.

\paragraph{Student modeling: reflection and help seeking.}
Student reflection and help-seeking behavior are widely associated with learning processes and self-regulated learning. We draw on the Reflective Thinking Questionnaire (RTQ)~\cite{kember2000rtq}, based on Mezirow's distinction between reflection and critical reflection~\cite{mezirow1991transformative}, together with discussion-board help-seeking behavior and self-efficacy measures~\cite{bandura2006guide}. In this work, these constructs serve as complementary evidence that student signals capture learning-relevant behavior rather than directly determining mediation priorities.

\paragraph{Knowledge-graph learner modeling.}
Knowledge-graph approaches represent relationships among concepts and learner understanding at the topic level. Prior work in A4L extracts curriculum knowledge graphs and topic-level learner representations~\cite{goel2025a4l}. We use topic-level gap prevalence derived from this pipeline as one source of evidence in the mediation process.

\paragraph{Human--AI co-agency and feedback coordination.}
AI teaching assistants such as Jill Watson demonstrate how AI systems can participate in classroom interactions~\cite{goel2018jill,taneja2024jill,kakar2024jill}. More recent work on bidirectional feedback and human--AI collaboration suggests that classroom support can benefit from integrating perspectives from multiple participants rather than relying on a single source of information~\cite{basu2025bidirectional,park2025human,park2026evaluating}. Our work explores how teacher, student, and researcher perspectives may be coordinated through a shared, transparent decision layer.

\section{Approach}
\label{sec:approach}

\begin{figure}[t]
\centering
\includegraphics[width=\linewidth-30pt]{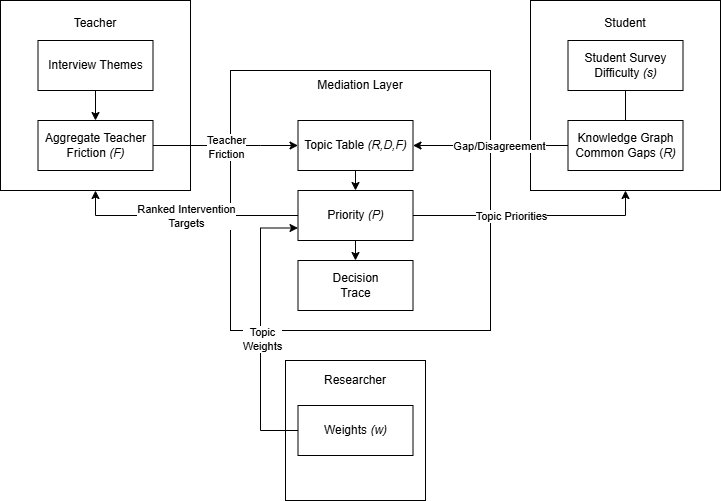}
\caption{Bidirectional feedback mediator. Teacher and student subsystems provide signals to a mediation layer that produces topic priorities and decision records. Researcher weights are specified rather than learned in the current implementation. Priorities are estimated without using graded outcomes.}
\label{fig:arch}
\end{figure}

Our approach organizes teacher, student, and researcher roles around a shared mediation layer that coordinates feedback across the classroom process (Figure~\ref{fig:arch}). Teacher signals derive from instructor use of a deployed Jill Watson analytics dashboard (Appendix~\ref{app:dashboard}): dashboard observations inform semi-structured interviews, which are coded into themes and summarized as teacher friction~$F$. Student signals derive from Jill Watson interaction traces and a mid-course survey, yielding gap prevalence~$R_t$, self-reported difficulty~$s_t$, and survey disagreement~$D_t$. Researcher-defined weights~$\mathbf{w}$ determine how these signals contribute to topic priorities and decision records.

\section{Methods}
\label{sec:methods}

\paragraph{Participants and data.}
Teacher data consist of five semi-structured interviews conducted following (IRB-approved) co-design and deployment of a Jill Watson analytics dashboard in a graduate-level CS course on AI \cite{goel2017using} at Georgia Institute of Technology \cite{goel2025a4l,thajchayapong2025evolution}. Student data consist of mid-course questionnaire responses ($n\!=\!279$) collected under standard learning analytics consent. Pseudonymized identifiers linked topics across data sources. Grades and demographic information were not used.

\subsection{Teacher, student, and researcher roles}

\paragraph{Teacher.}
Interview transcripts were segmented and coded using a 12-theme codebook (Appendix~\ref{app:teacher}). Coded themes were summarized into teacher signals used by the mediation layer. Teacher friction $F$ was computed as the proportion of usability-friction and trust-uncertainty themes among all coded segments ($F\!=\!0.220$ for this offering).

\paragraph{Student.}
Jill Watson questions were mapped to course topics and analyzed for potential learning difficulties, which were aggregated into cohort-level gap prevalence $R_t$. Survey responses mapped to the same topics and provided self-reported difficulty $s_t$ (Appendices~\ref{app:student} and~\ref{app:survey}). RTQ, help-seeking, and self-efficacy measures were used only for validation and did not contribute directly to the priority score $P_t$.

\paragraph{Researcher.}
Weights $\mathbf{w}=(0.70,0.20,0.10)$ reflected priorities established during co-design sessions: greater emphasis was placed on student learning difficulties, followed by disagreement signals and teacher concerns. Alternative weights produced similar rankings (Appendix~\ref{app:weights}).

\paragraph{Mediation.}
Survey disagreement was computed as

\begin{equation}
\label{eq:d_t}
D_t=\bigl|R_t-s_t\bigr|
\end{equation}

when survey difficulty $s_t$ was available; otherwise $D_t=0$. Topic priority was then computed as

\begin{equation}
\label{eq:p_t}
P_t \;=\; w_rR_t+w_dD_t+w_fF
\end{equation}

where higher values indicate greater instructional priority. Topics were ranked by $P_t$, and the mediation layer produced a topic table together with a per-topic decision record (Appendix~\ref{app:trace}).

\section{Validation}
\label{sec:validation}
Because the mediation layer is designed to support decision-making rather than predict outcomes, we evaluated whether ranked topic priorities aligned with independent evidence from instructors and students. We report three complementary checks from the same course offering; these are intended as descriptive and exploratory evidence rather than predictive performance claims.

\paragraph{Instructor agreement.}
Five interviews yielded 41 coded segments mapped to topics through a theme--topic table (Appendix~\ref{app:teacher}). Among the top five mediated topics, three overlapped with instructor-identified concern topics (e.g., analogical reasoning, planning, knowledge-based AI). Rank agreement between mediated priorities and instructor concern levels was high (Spearman $\rho=0.80$, $p<.001$, $n=19$ topics with survey data). Given the small number of interviews, we interpret this as descriptive evidence that the mediator surfaces topics instructors already consider important.

\paragraph{Student alignment.}
Self-reported topic difficulty $s_t$ (derived from Q5 topic items) provided an independent student-side comparison. Topic priority $P_t$ was moderately associated with self-reported difficulty ($\rho=0.46$, $p=.048$, $n=19$), compared with a weaker association for gap prevalence alone ($\rho=0.29$, $p=.23$). This suggests that higher-priority topics align more closely with learner-reported difficulties without relying on grades or outcome labels.

\paragraph{Multi-signal integration.}
The disagreement signal $D_t=|R_t-s_t|$ increased priority when observed learning difficulties and student self-reports differed. Removing this term left the top-five topics unchanged but reduced agreement with the default ranking ($\rho=0.96$; Appendix~\ref{app:weights}). For isolated learners, students not identified by any individual signal source, exposure to higher-priority topics during weeks they reported difficulty ranked them above comparison students (AUC $=0.96$, $n_{\mathrm{iso}}=2$, $n_{\mathrm{comp}}=31$, 95\% CI $[0.89,1.00]$, permutation $p=.011$), compared with AUC $=0.91$ for gap prevalence alone. While exploratory, these findings suggest that combining multiple signals may reveal support needs not visible from any individual source.

\section{Results}
\label{sec:results}

\subsection{Topic priorities and rank stability (H1)}
Teacher interview coding produced a global teacher friction score of $F=0.220$. The course knowledge graph contained 54 topic nodes and 47 prerequisite links. Combined with gap prevalence ($R_t$) and survey disagreement ($D_t$), these signals produced ranked topic priorities and decision records (Appendix~\ref{app:trace}). Using default weights $(w_r,w_d,w_f)=(0.70,0.20,0.10)$, the highest-ranked topics were analogical reasoning, frames, planning, case-based reasoning, and knowledge-based AI. Alternative weight settings produced similar rankings and preserved top-topic overlap (Appendix~\ref{app:weights}), suggesting that priorities remained relatively stable across reasonable weighting choices.

\subsection{Student behavioral signals (H2)}

\begin{table}[h]
\centering
\small
\caption{Student behavioral measures associated with topic understanding ($n=279$). Additional analyses appear in Appendix~\ref{app:h2_extended}.}
\label{tab:student}
\begin{tabular}{llll}
\toprule
\textbf{Construct} & \textbf{Relationship} & \textbf{Stat} & \textbf{$p$} \\
\midrule
Self-efficacy & Q13 $\rightarrow$ Q5 & $r=0.427$ & $<.001$ \\
RTQ & Understanding $\rightarrow$ Q5 & $r=0.369$ & $<.001$ \\
Help-seeking & Ed $\rightarrow$ Q5 & $r=0.168$ & $.005$ \\
Help-seeking & Found vs.\ not-found & $d=0.479$ & --- \\
\bottomrule
\end{tabular}
\end{table}

Self-efficacy showed the strongest association with topic understanding ($r=0.427$), followed by RTQ Understanding ($r=0.369$) and help-seeking behavior ($r=0.168$). The contrast between students who found answers and those who did not suggested a moderate effect size ($d=0.479$). These results provide complementary evidence that reflection and help-seeking measures capture learning-related behavior.

\subsection{Learners surfaced through multi-signal integration (H3)}

Three risk channels (understanding, help-seeking, and reflection) were normalized and combined:

\begin{equation}
\label{eq:sigma}
\sigma
=
0.35\,\rho_{\text{raw}}
+
0.35\,\rho_{\text{help}}
+
0.30\,\rho_{\text{refl}}
\end{equation}

Students identified by $\sigma$ while remaining undetected by individual signals were considered \emph{isolated learners} (details in Appendix~\ref{app:isolated_mechanics}). A representative learner (Case~A) showed no single-source concern signal: Q5 mean $=5.08/6$, RTQ-UND $=4.00$, and one unresolved help event, yet produced $(\rho_{\text{raw}}, \rho_{\text{help}}, \rho_{\text{refl}}) = (0.31, 0.70, 0.52)$ and $\sigma=0.509$. Across the full cohort ($n=279$), three students ($2.2\%$ of 134 baseline-unidentified students) were surfaced through synthesis alone, while 22 additional students were identified jointly by synthesis and at least one individual signal. One learner reported understanding throughout the course and would not have been identified through topic understanding measures alone; their only observable concern signal was unresolved help-seeking behavior. These findings suggest that combining multiple signals may reveal support needs that are difficult to identify through any single source.

\section{Discussion}
Our findings provide preliminary support for H1--H3: (H1) topic priorities remained relatively stable across weighting choices, suggesting that recommendations were not highly sensitive to small changes in assumptions, (H2) reflection, help-seeking, and self-efficacy aligned with learning-related constructs, providing complementary evidence that student signals capture meaningful aspects of learning behavior, and (H3) multi-signal integration surfaced learners not identified through individual signal sources alone. Our findings suggest that a transparent coordination process may help organize multiple sources of classroom feedback when graded outcomes are unavailable. Rather than relying on a single signal source, the mediation layer combines teacher concerns, learner difficulties, and disagreement between observed and self-reported evidence into shared topic priorities. Case~A (Section~\ref{sec:results}) illustrates this behavior: a learner not identified through any individual signal source became visible only when multiple signals were considered together. Relative to teacher reflection dashboards~\cite{cai2025aipoweredreflection}, this work explores a broader coordination role for AI support. Rather than focusing on instructor reflection alone, the mediation process organizes teacher, student, and researcher perspectives around shared decision records that can be inspected and reviewed.

\paragraph{Reproducibility and future directions.}
We provide a lightweight reference implementation of the mediation layer (\SupplementRepo) for replication and exploratory use. The implementation returns topic priorities and decision records. We present it as a demonstration artifact. Future work may investigate richer classroom signals and human oversight workflows for semi-automated educational agents.

\paragraph{Limitations.}
This study reflects one graduate-level course offering. Instructor agreement is based on five interviews, survey measures are cross-sectional, and assignment artifacts were not linked at the individual level. Findings should therefore be interpreted as preliminary and exploratory rather than population-level estimates.

\section{Conclusion}
AI-augmented classrooms generate multiple forms of teacher and student feedback before graded outcomes become available, yet these signals can be difficult to translate into timely instructional decisions. This work explored an interpretable decision layer that combines teacher concerns and learner signals into shared topic priorities with accompanying decision records. Preliminary evidence from one graduate course offering suggested alignment with instructor concerns and learner-reported difficulty, while multi-signal integration surfaced learners not identified through individual signal sources alone. Rather than replacing instructor judgment, we view the mediation layer as a mechanism for supporting transparent coordination among teacher, student, and researcher perspectives. We consider this work an initial step toward understanding how human--AI co-agency may support classroom decision-making when feedback is incomplete.

\begin{acknowledgments}
We thank members of the A4L team in Georgia Tech's Design Intelligence Lab (DILab; dilab.gatech.edu) for their contributions to this research. This research is supported by a US National Science Foundation grant (Grant No. 2247790) to the National AI Institute for Adult Learning and Online Education (AI-ALOE; aialoe.org).
\end{acknowledgments}

\section*{Declaration on Generative AI}
  During the preparation of this work, the author(s) used ChatGPT, Claude in order to: Draft content, Paraphrase and reword. After using this tool/service, the author(s) reviewed and edited the content as needed and take(s) full responsibility for the publication’s content.
  \newline

\bibliography{refs}

\section{Appendices}
\label{sec:appendices}

\appendix

\section{Dashboard interface and teacher workflow}
\label{app:dashboard}

\begin{figure}[h]
\centering
\includegraphics[width=\linewidth-30pt]{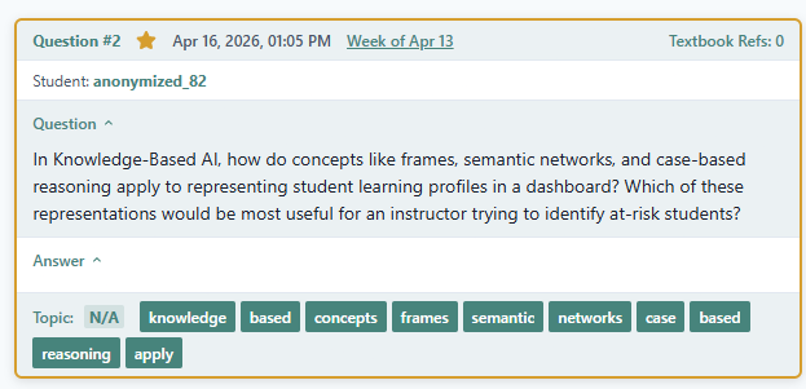}
\caption{Deployed Jill Watson dashboard Q\&A audit view used during the course offering. Instructors reviewed unresolved student questions and reflected on instructional concerns that later informed coded teacher signals. Identifiers are anonymized.}
\label{fig:teacher_dashboard}
\end{figure}

Teacher signals were collected through interaction with a deployed Jill Watson analytics dashboard. Figure~\ref{fig:teacher_dashboard} shows the Q\&A audit interface used by instructors to review unresolved student questions requiring additional attention. Semi-structured interviews following dashboard-informed reflection were coded into themes that supplied teacher signals to the mediation layer.

\section{Knowledge graph details}
\label{app:kg}

The course knowledge graph contains 54 topic nodes and 47 prerequisite relationships extracted from course materials and textbook structure. Topics represent conceptual units in the KBAI curriculum (e.g., analogical reasoning, planning, frames, and case-based reasoning). The graph was used to map Jill Watson questions to topics and derive cohort-level gap prevalence $R_t$, which served as input to the mediation layer.

\section{Student gap pipeline}
\label{app:student}
Student questions from Jill Watson were mapped to one or more course topics and classified as \textit{correct}, \textit{incorrect}, or \textit{unknown} using an ensemble text classifier (80.0\% accuracy on a human-annotated gold set, $n=70$). Topic-level gaps were identified from question patterns and aggregated into cohort prevalence $R_t$, defined as the fraction of learners showing trace evidence of difficulty on topic~$t$ ($|\mathcal{S}|=159$). Mid-course survey difficulty $s_t$ was joined using shared topic identifiers (Appendix~\ref{app:survey}). RTQ, help-seeking, and self-efficacy measures were used only for H2 validation and did not contribute directly to $P_t$.

\section{Survey instrument}
\label{app:survey}
The mid-course questionnaire (\emph{JW Dashboard Questionnaire, 1st Half}; $n=279$) included:

\begin{itemize}
\item \textbf{Topic understanding (Q5):} topic-level self-ratings used to derive survey difficulty $s_t$;
\item \textbf{Help-seeking (Q6--Q11):} uncertainty and discussion-search behavior;
\item \textbf{Reflective Thinking Questionnaire (RTQ):} Understanding, Reflection, and related subscales~\cite{kember2000rtq};
\item \textbf{Self-efficacy (Q13):} domain-specific capability beliefs~\cite{bandura2006guide}.
\end{itemize}

Item counts appear in Table~\ref{tab:survey_map}; full wording is available in the supplement repository (Appendix~\ref{app:repro}).

\section{Extended student empirical outcomes (H2)}
\label{app:h2_extended}

Table~\ref{tab:student_extended} reports additional student behavioral relationships used as complementary evidence for H2. These measures were used to assess whether reflection, help-seeking, and self-efficacy align with learning-related constructs; they were not included in the mediation score. Non-significant relationships are reported for completeness. Table~\ref{tab:survey_map} summarizes the questionnaire components used throughout the study.

\begin{table}[h]
\centering
\small
\caption{Full student empirical outcomes ($n=279$). Reported as complementary evidence for H2; not integrated with the mediation score.}
\label{tab:student_extended}
\begin{tabular}{llll}
\toprule
\textbf{Construct} &
\textbf{Relationship} &
\textbf{Statistic} &
\textbf{$p$}\\
\midrule
Self-efficacy & Q13 $\rightarrow$ Q5 & $r=0.427$ & $<.001$ \\
RTQ & Understanding $\rightarrow$ Q5 & $r=0.369$ & $<.001$ \\
RTQ & Critical Refl.\ $\rightarrow$ Q5 & $r=0.077$ & $.201$ \\
Help-seeking & Ed $\rightarrow$ Q5 (Pearson) & $r=0.168$ & $.005$ \\
\bottomrule
\end{tabular}
\end{table}

\begin{table}[h]
\centering
\small
\caption{Survey item map ($n=279$).}
\label{tab:survey_map}
\begin{tabular}{llc}
\toprule
\textbf{ID} & \textbf{Construct} & \textbf{\# Items} \\
\midrule
Q5 & Topic understanding & 13 \\
Q6--Q11 & Help-seeking behavior & 6 \\
Q13 & Self-efficacy & 7 \\
RTQ & Reflective Thinking & 16 \\
Q14 & Open feedback & 1 \\
\bottomrule
\end{tabular}
\end{table}

\section{Teacher subsystem}
\label{app:teacher}

Teacher interview coding used a 12-theme codebook including instructional insights, usability concerns, trust and verification, and missing-context themes. Teacher friction $F$ summarizes concern signals entering the mediation layer and is computed as:

\[
F=
\frac{
\#\text{friction themes}
}{
\#\text{all themes}
}
=
0.220
\]

\section{Decision record examples}
\label{app:trace}

Table~\ref{tab:decision_examples} illustrates two representative outputs from the mediation layer. Panel~A shows how teacher and student signals are translated into topic priorities that can guide instructional attention. Panel~B shows how integrating multiple signals can surface learners who may otherwise remain unnoticed.

\begin{table}[h]
\centering
\small
\caption{Representative mediation outputs. Panel A: topic priorities for instructional attention. Panel B: learner synthesis for identifying potentially overlooked learners.}
\label{tab:decision_examples}

\textbf{(A) Topic priority record}

\vspace{0.05cm}

\begin{tabular}{lrrrrr}
\toprule
\textbf{Topic} & $R_t$ & $D_t$ & $F$ & $P_t$ & Rank \\
\midrule
Analogical reasoning & .157 & .036 & .220 & .139 & 1 \\
Planning             & .101 & .036 & .220 & .100 & 3 \\
\bottomrule
\end{tabular}

\vspace{0.15cm}

\textbf{(B) Learner synthesis contrast}

\vspace{0.05cm}

\begin{tabular}{lrrrrl}
\toprule
\textbf{Learner} &
$\rho_{\text{raw}}$ &
$\rho_{\text{help}}$ &
$\rho_{\text{refl}}$ &
$\sigma$ &
\textbf{Isolated?}\\
\midrule
case\_a           & .31 & .70 & .52 & .509 & Yes\\
baseline\_flagged & .90 & .20 & .15 & .430 & No\\
\bottomrule
\end{tabular}
\end{table}

Panel~A illustrates how multiple signals contribute to topic-level recommendations that can help prioritize instructional support. Panel~B illustrates a learner identified only after combining multiple signals: \texttt{case\_a} exceeds the synthesis threshold despite no individual signal source indicating concern.

\section{Weight sensitivity}
\label{app:weights}

The mediation layer combines teacher and student signals using researcher-defined weights. Because these choices reflect assumptions rather than learned parameters, we examined whether reasonable alternative settings produced substantially different recommendations. Table~\ref{tab:weights} shows representative weight profiles.

\begin{table}[h]
\centering
\small
\caption{Representative weight profiles.}
\label{tab:weights}
\begin{tabular}{lccc}
\toprule
\textbf{Profile} & $w_r$ & $w_d$ & $w_f$\\
\midrule
Default & 0.70 & 0.20 & 0.10\\
Higher disagreement & 0.60 & 0.40 & 0.00\\
\bottomrule
\end{tabular}
\end{table}

Recommendations remained relatively stable across profiles: the top-10 topics completely overlapped between profiles (10/10 overlap), and removing disagreement signals left the top five topics unchanged (Spearman $\rho=0.96$). These results suggest that small changes in assumptions did not substantially alter the recommendations presented to instructors.

\section{Supplement repository}
\label{app:repro}

A reference implementation for replication and exploratory use is available at \SupplementRepo. It accepts classroom-level observations and returns topic priorities, decision records, and recommendations.

\section{Isolated-learner channel mechanics (H3)}
\label{app:isolated_mechanics}

The isolated-learner synthesis process combines three normalized risk channels, shown below, where larger values indicate greater concern:

\begin{itemize}
\item \textbf{Understanding} ($\rho_{\text{raw}}$): inverse of mean topic-understanding scores;
\item \textbf{Help-seeking} ($\rho_{\text{help}}$): unresolved or absent discussion help behavior;
\item \textbf{Reflection} ($\rho_{\text{refl}}$): inverse of RTQ Understanding and Reflection measures.
\end{itemize}

The synthesis score $\sigma$ (Eq.~\ref{eq:sigma}) combines these channels into a single concern score. Students in the upper half of the synthesized risk distribution ($\sigma\ge0.50$) were considered higher concern. An \emph{isolated learner} is a student surfaced by $\sigma$ while remaining unidentified by any individual signal source.

\end{document}